\documentclass{article}

\usepackage[preprint]{neurips_2026}

\usepackage[utf8]{inputenc} %
\usepackage[T1]{fontenc}    %
\usepackage{hyperref}       %
\usepackage{url}            %
\usepackage{booktabs}       %
\usepackage{amsfonts}       %
\usepackage{nicefrac}       %
\usepackage{microtype}      %
\usepackage{xcolor}         %
\usepackage{graphicx}
\usepackage{bbm}
\usepackage{multirow}
\usepackage{bm}
\usepackage[table]{xcolor}
\usepackage{amsmath}
\usepackage{pifont}
\usepackage{wrapfig}  %
\usepackage{arydshln}
\usepackage{enumitem}
\title{GC-MoE: Genomics-Guided Cell-Type-Specific Mixture of Experts for Histology-Based \\ Single-Cell Spatial Transcriptomics}

\author{
Kaito Shiku$^{1,2}$ \quad
Ahtisham Fazeel Abbasi$^{2,3}$ \quad
Ryoma Bise$^{1}$ \quad
Yuichiro Iwashita$^{2,3,6}$ \\
\textbf{Kazuya Nishimura}$^{4}$ \quad
\textbf{Andreas Dengel}$^{2,3,5}$ \quad
\textbf{Muhammad Nabeel Asim}$^{2,5,6}$ \
\\
$^{1}$ Kyushu University, Japan \\
$^{2}$ German Research Center for Artificial Intelligence (DFKI GmbH), Germany \\
$^{3}$ RPTU University Kaiserslautern-Landau, Germany \\
$^{4}$ The University of Osaka, Japan \\
$^{5}$ IntelligentX GmbH (intelligentx.com), Germany \\
$^{6}$ Osaka Metropolitan University, Japan \\
\texttt{kaito.shiku@human.ait.kyushu-u.ac.jp} \
}

\begin{document}

\maketitle

\begin{abstract}
Histology-based single-cell spatial transcriptomics (ST) estimation aims to predict gene expression for individual cells from histopathological images and cell locations, reducing the need for costly single-cell ST measurements. Unlike existing histology-to-ST methods that mainly predict spot-level profiles for local regions containing multiple cells, this task requires modeling cell-to-cell expression variability, which is strongly structured by cell type. We propose Genomics-Guided Cell-Type-Specific Mixture-of-Experts (GC-MoE), which estimates cell-type probabilities with a routing network and softly combines cell-type-specific experts for gene expression prediction. To further encode cell-type-dependent gene programs, we introduce the Cell-Type-Specific Co-Expression-Aware Predictor (CAP), together with a lightweight Cell-to-Cell Interaction Attention (C2CA) module for neighboring-cell context. Experiments and ablations on public single-cell ST datasets show consistent improvements over existing single-cell and adapted spot-level baselines.
\end{abstract}

\section{Introduction}
Histology-based spatial transcriptomics (ST) prediction aims to infer spatially resolved gene expression from routinely available hematoxylin and eosin (H\&E) images, and existing studies have largely focused on spot-level measurements~\cite{he2020integrating,pang2021leveraging,xie2023spatially,chung2024accurate,ganguly2025merge,yang2023exemplar, yang2024spatial}. ST measures the abundances of many genes at spatially indexed locations in tissue, providing critical insights into disease progression, drug response, and tissue function~\cite{marx2021method,staahl2016visualization}. In spot-level ST, each capture spot corresponds to a local tissue region, typically containing multiple cells, that can be matched to a patch in the H\&E image. For each spot, ST provides a gene expression vector containing the measured abundances of many genes in that local region. Accordingly, existing methods learn a mapping from the H\&E image patch around each spot to its spot-level expression vector~\cite{he2020integrating,pang2021leveraging,chung2024accurate}. This setting is attractive because it can computationally augment routine or archival histology images with molecular information, enabling large-scale or retrospective tissue analysis without requiring ST measurements for every specimen.

Recent advances in ST technologies now enable gene expression measurements at single-cell resolution, creating growing demand for cell-resolved molecular maps~\cite{janesick2023high, oliveira2025high}. Compared with spot-level measurements, single-cell ST can distinguish molecular programs of neighboring cell populations, identify rare cell populations, and characterize spatial heterogeneity within tissue microenvironments. However, acquiring single-cell ST remains costly and technically demanding, limiting its use at large scale. This motivates histology-based single-cell ST estimation, which aims to infer cell-resolved gene expression from histology images and detected cell locations.

\begin{figure}[t]
      \centering
        \includegraphics[width=1.0\linewidth]{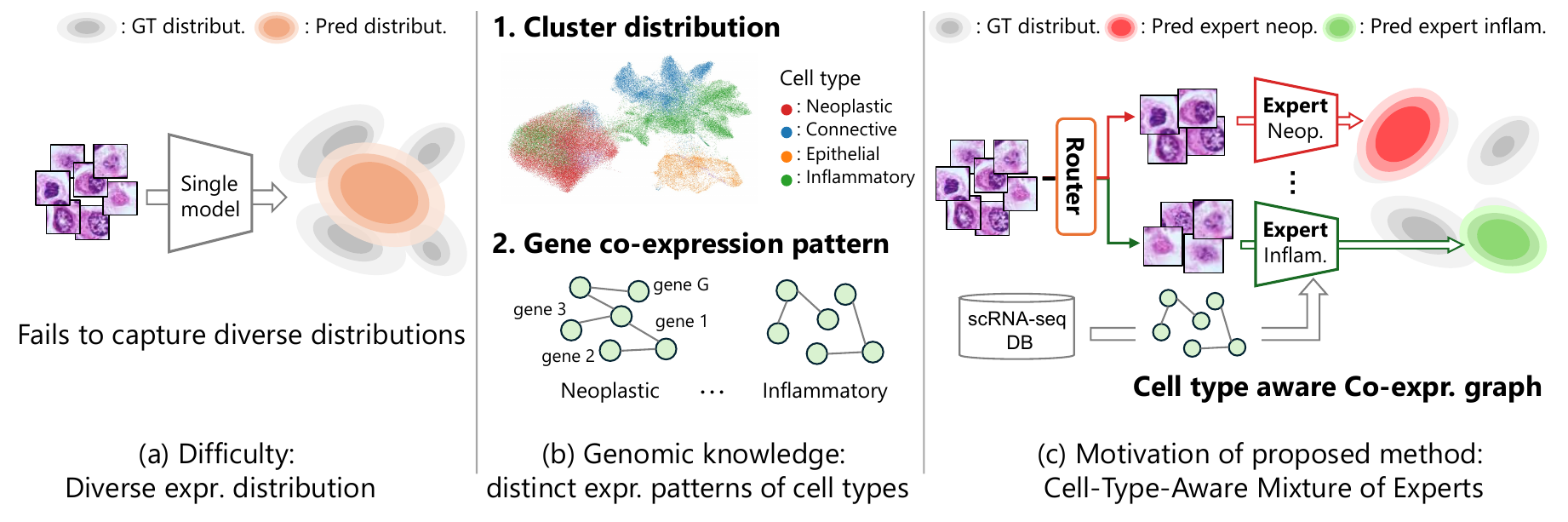}
        \caption{(a) Difficulty of Single-Cell ST Estimation: Despite subtle variations in single-cell morphology, gene expression exhibits highly diverse patterns. (b) Genomic knowledge: Single-cell gene expression is known to exhibit distinct patterns across different cell types. (c) Motivation of proposed method: Mixture-of-Experts Framework Using Cell-Type-Specific Expert Models.} 
        \vspace{-3mm}
        \label{fig:intro}
\end{figure}

At single-cell resolution, the model must predict differences among individual cells rather than only the molecular composition of a local tissue region. In spot-level ST, each target expression vector is measured from a capture spot and therefore pools signals from multiple cells, making cell-to-cell differences less explicit in the supervision signal. In single-cell ST, each cell has its own target expression vector, and the variation among neighboring or morphologically similar cells becomes the direct target of prediction. As illustrated in Figure~\ref{fig:intro}(a), when such diverse cell-level targets are modeled with a single shared predictor, the predictions tend to move toward intermediate expression profiles, failing to capture separated modes in the true expression distribution.

Cell type provides a natural structure for organizing single-cell expression variability. Single-cell expression profiles vary substantially across cell types, with distinct marker genes and transcriptional programs~\cite{schaum2018single}. As shown in Figure~\ref{fig:intro}(b), cells of different types form distinct regions in expression space, while cells of the same type tend to share characteristic expression patterns. In addition to this cell-level structure, gene expression within each cell type is also organized by gene-gene relationships: genes participating in the same biological program often exhibit coordinated co-expression, and such co-expression patterns can differ across cell types~\cite{su2023cell}. This suggests that single-cell ST prediction should preserve both cell-type-specific expression modes across cells and gene-level structure within each mode, rather than treating all cells as samples from one homogeneous regression problem.

Motivated by this perspective, we propose \textbf{Genomics-Guided Cell-Type-Specific Mixture of Experts (GC-MoE)} for histology-based single-cell ST estimation. As illustrated in Figure~\ref{fig:intro}(c), GC-MoE uses a routing network to estimate cell-type probabilities for each target cell and softly combines cell-type-specific experts, allowing the model to represent multiple cell-type-dependent transcriptional programs without assuming that cell type is observed at inference time. To align each expert with the gene program of its corresponding cell type, we introduce the Cell-Type-Specific Co-Expression-Aware Predictor (CAP), which constructs a separate gene co-expression graph for each cell type from external single-cell RNA-sequencing (scRNA-seq) data~\cite{barrett2012ncbi} and uses it to guide the corresponding expert. We additionally include a lightweight Cell-to-Cell Interaction Attention (C2CA) module that refines predictions using neighboring-cell features as local contextual cues.

We evaluate GC-MoE on multiple publicly available single-cell ST datasets against an existing single-cell ST prediction method and spot-level ST prediction methods adapted to cell-level inputs. GC-MoE achieves the best performance across all datasets, demonstrating the benefit of explicitly modeling cell-type-structured expression variability. Additional analyses show that GC-MoE better recovers cell-type-specific expression distributions, learns gene predictors aligned with co-expression relationships, and produces spatial expression maps closer to ground-truth measurements.

Our contributions are summarized as follows:
\begin{itemize}
\item We formulate histology-based single-cell ST estimation as modeling cell-type-structured expression variability that becomes explicit at single-cell resolution.
\item We propose GC-MoE, a genomics-guided mixture-of-experts framework that softly combines cell-type-specific experts through a routing network, enabling the model to capture multiple cell-type-dependent expression modes without requiring observed cell types at inference time.
\item We introduce CAP, which incorporates cell-type-specific gene co-expression priors from external scRNA-seq data into each expert, and include C2CA as a lightweight refinement module using neighboring-cell context.
\item We validate GC-MoE on multiple public single-cell ST datasets, showing improvements over existing single-cell and adapted spot-level baselines together with analyses of expression distributions, co-expression-aware predictors, and spatial expression patterns.
\end{itemize}

\section{Related Work}

\noindent
{\bf Spatial Transcriptomics Estimation  for Pathological
Images.}
Recently, gene expression estimation from histopathology images has attracted increasing attention.
Most existing approaches focus on predicting gene expression at the spot level, where each spot contains multiple cells~\cite{he2020integrating, pang2021leveraging, xie2023spatially, chung2024accurate, ganguly2025merge, yang2023exemplar, yang2024spatial}.
ST-Net~\cite{he2020integrating} predicts spot-level te gene expression from image features using a multi-output regression head, enabling simultaneous prediction of multiple genes.
Recent state-of-the-art methods focus on modeling spatial correlations within tissues, including Vision Transformer-based approaches such as HisToGene~\cite{pang2021leveraging} and TRIPLEX~\cite{chung2024accurate}, as well as graph neural network (GNN) based methods~\cite{ganguly2025merge, yang2023exemplar, yang2024spatial}.
In a different direction, BLEEP~\cite{xie2023spatially} learns a shared latent space between histology images and gene expression by aligning image features with gene expression via contrastive learning, inspired by frameworks such as CLIP~\cite{radford2021learning}.

While extensive efforts have focused on spot-level gene expression estimation, single-cell-level studies remain underexplored. 
To date, GHIST~\cite{fu2025spatial} is one of the few pioneering works on single-cell ST estimation.
GHIST adopts a multi-task learning framework with a U-Net~\cite{ronneberger2015u} backbone for joint cell segmentation and gene expression prediction. 
However, it employs a single model across cell types, making it difficult to capture diverse expression patterns in single-cell ST.

\noindent
{\bf Super-Resolution of Spatial Transcriptomics from Spot-Level to Single-Cell Resolution.}
Super-resolution approaches aim to address the limitation of conventional ST technologies, which can only measure gene expression at the spot level containing multiple cells, by reconstructing single-cell ST from spot-level observations~\cite{xue2025inferring,li2024high,hu2023deciphering}. While these methods share a similar goal with ours, they fundamentally differ in that they assume access to costly spot-level ST measurements even at inference time to guide the reconstruction.
In contrast, our single-cell ST estimation task considers a more practical setting in which only histological images are required at inference time and no additional ST measurements are needed. As a result, existing super-resolution approaches cannot be directly applied to our problem.

\noindent
{\bf Mixture of Experts (MoE).}
MoE has recently gained significant attention in deep learning~\cite{huai2025cl, wei2025mixture, li2025moe, lu2026segmote, kim2025mixture}. It is a simple yet effective framework in which multiple models, referred to as experts, are trained to specialize in different tasks or data characteristics.
Originally, MoE was proposed in the field of large language modeling. In recent years, it has been widely applied to various computer vision tasks, including visual question answering (VQA)~\cite{huai2025cl}, segmentation~\cite{wei2025mixture, li2025moe, lu2026segmote}, multitask learning~\cite{chen2023mod}, and domain adaptation~\cite{kim2025mixture}.

To the best of our knowledge, no prior studies in ST estimation have introduced expert models by explicitly leveraging the fact that gene expression exhibits distinct patterns across different cell types.

\section{Histology-Based Single-Cell Spatial Transcriptomics Estimation}
\subsection{Problem Setting}
We study histology-based single-cell ST estimation, where the goal is to predict a gene expression profile for each target cell in an H\&E tissue image. Given a local H\&E image patch $\mathbf{x}$ and the locations of $N$ target cells $\mathcal{P}=\{\mathbf{p}_i\}_{i=1}^{N}$, the model predicts a gene expression vector $\hat{\mathbf{y}}_i \in \mathbb{R}^{G}$ for each cell, where $G$ denotes the number of target genes.

For training, the corresponding ground-truth expression vector $\mathbf{y}_i \in \mathbb{R}^{G}$ is available for each target cell. In typical single-cell ST datasets, cell-level expression vectors are constructed by aggregating gene-wise spatial expression maps within cell-instance masks. Such masks can be obtained automatically from a DAPI-stained image when available, but the proposed expression predictor only requires the H\&E image and target-cell locations at inference time.

\begin{figure}[t]
      \centering
        \includegraphics[width=0.95\linewidth]{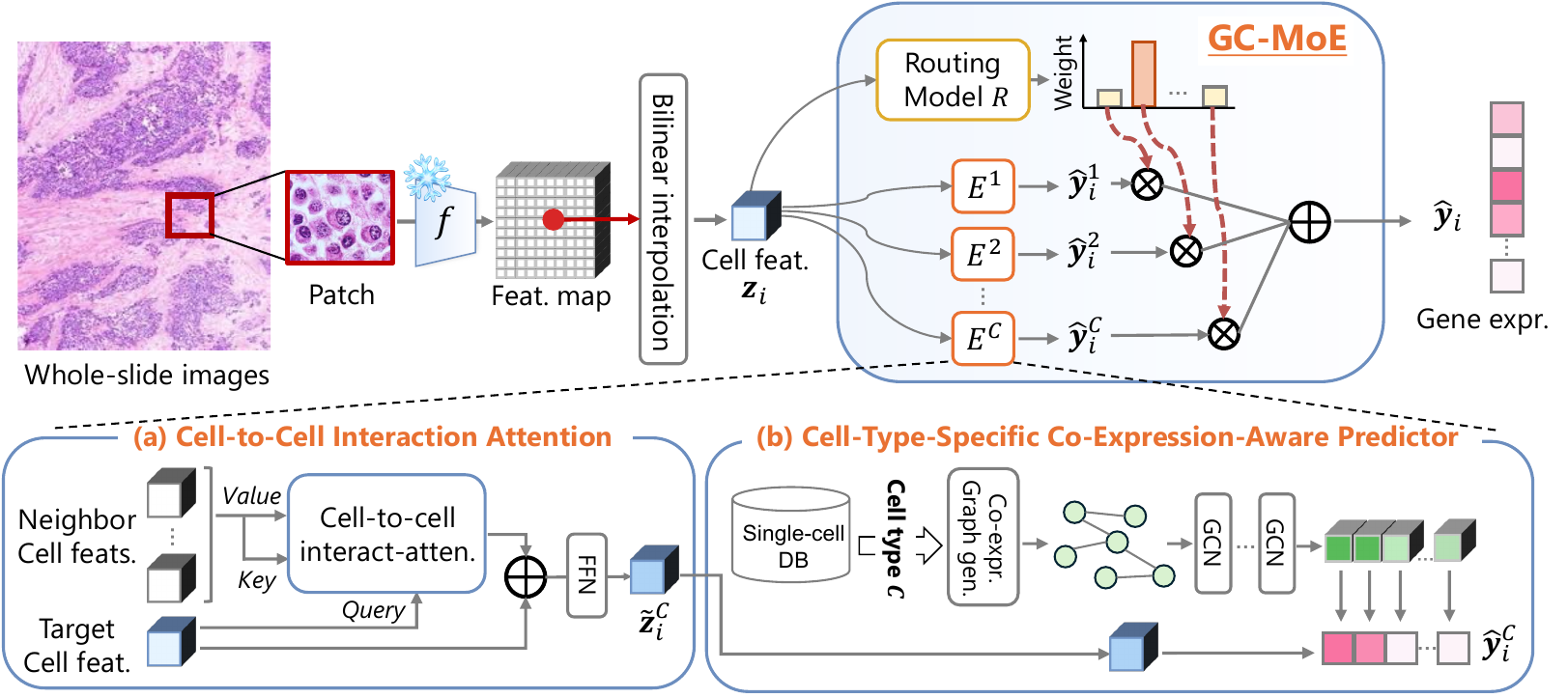}
        \caption{\textbf{Overview of proposed Genomics-Guided Cell-Type-Specific Mixture of Experts (GC-MoE).} GC-MoE consists of cell-type-specific expert models and a routing model that dynamically assigns experts to each target cell. Each expert further incorporates two modules to enhance specialization: a Cell-to-Cell Interaction Attention (C2CA) module and a Cell-Type-Specific Co-Expression-Aware Predictor (CAP) module.}
        \label{fig:method}
    \vspace{-4mm}
\end{figure}

\subsection{Genomic Knowledge-Guided Cell-Type-Specific Mixture of Experts}
\subsubsection{Overview}
Figure~\ref{fig:method} shows an overview of the proposed \textbf{Genomics-Guided Cell-Type-Specific Mixture of Experts (GC-MoE)}. 
Given a histological image patch and target-cell positions, GC-MoE first obtains cell-centered image features by sampling a patch-level feature map at the target-cell positions, and then predicts gene expression by softly combining a set of cell-type-specific expert models $\{E_j\}_{j=1}^{C}$ through a routing model $R$.
Each expert is specialized for a particular cell type, while the routing model estimates cell-type probabilities for each target cell from its image feature.
During training, the routing model is supervised using cell-type labels $c_i \in {1,\dots,C}$ automatically annotated from measured single-cell expression profiles by an existing cell-type annotation method~\cite{nishimura2026leveraging}; these labels are not required during inference.
Each expert further incorporates CAP, which leverages cell-type-specific gene co-expression patterns from external scRNA-seq databases, and C2CA, which uses neighboring-cell features as local context.
In the following sections, we describe cell-centered image feature extraction, expert assignment by the routing model, and the expert modules in detail.

\subsubsection{Cell-Centered Image Feature Extraction}
To obtain image features for target cells while retaining local tissue context, we first encode the entire histological image patch using a pathology foundation model. 
Given a histological image patch $\mathbf{x}$, the image encoder produces a feature map $\mathbf{Z} = f(\mathbf{x}) \in \mathbb{R}^{H \times W \times D}$, where $H$ and $W$ denote the spatial resolution of the feature map, and $D$ denotes the feature dimension.

For the $i$-th target cell, we sample a feature vector from $\mathbf{Z}$ at its center position $\mathbf{p}_i$. 
After mapping $\mathbf{p}_i$ to the feature-map coordinate system, we apply bilinear interpolation to obtain a continuous cell-centered feature:
\begin{equation}
\mathbf{z}_i = \mathrm{BilinearInterpolation}(\mathbf{Z}, \mathbf{p}_i), \quad \mathbf{z}_i \in \mathbb{R}^{D}.
\end{equation}
Applying this operation to all target cells yields the cell-level feature set $\mathcal{Z}=\{\mathbf{z}_i\}_{i=1}^{N}$, which is used as the input to GC-MoE.

\subsubsection{Dynamic Expert Assignment in Cell-Type-Aware Mixture-of-Experts}
We dynamically assign cell-type-specific expert models to each target cell based on its image features. Specifically, expert assignment is achieved by aggregating the predictions of all expert models using weights estimated by a routing model $R$.

Given a cell-level feature $\mathbf{z}_i$, the $j$-th expert predicts the gene expression as follows:
$\hat{\mathbf{y}}_i^{j} = E^j(\mathbf{z}_i)$.
In parallel, a routing model $R$ takes the same feature $\mathbf{z}_i$ as input and estimates the assignment weights over experts by applying a softmax function to its output:
$\boldsymbol{\alpha}_i = \mathrm{softmax}(R(\mathbf{z}_i)) \in \mathbb{R}^{C}$.

The final prediction is obtained by aggregating the outputs of all expert models using the estimated probabilities as weights:
\begin{equation}
\hat{\mathbf{y}}_i = \sum_{j=1}^{C} \alpha_i^{j} \, \hat{\mathbf{y}}_i^{j}.
\end{equation}

In this process, by treating the routing weights as soft probabilities, the model can flexibly combine multiple experts while enabling end-to-end training of the routing model.

For training this cell-type-aware routing mechanism, we introduce an expression prediction loss and a routing loss. 
For a mini-batch $\mathcal{B}$, they are defined as
\begin{equation}
\mathcal{L}_{\mathrm{pcc}} =
1-\frac{1}{G}\sum_{g=1}^{G}
\mathrm{PCC}\left(\{\hat{y}_{i,g}\}_{i\in\mathcal{B}},\{y_{i,g}\}_{i\in\mathcal{B}}\right),
\quad
\mathcal{L}_{\mathrm{route}} =
\frac{1}{|\mathcal{B}|}\sum_{i\in\mathcal{B}}\mathrm{CE}(\boldsymbol{\alpha}_i,c_i).
\label{eq:loss_pcc_route}
\end{equation}
Here, $\mathcal{L}_{\mathrm{pcc}}$ measures gene-wise agreement between predicted and ground-truth expression across cells, while $\mathcal{L}_{\mathrm{route}}$ encourages the routing probabilities to match the training-only cell-type labels. 
The full training objective, including the co-expression regularization, is described in Section~\ref{sec:overall_loss}.
\subsubsection{Cell-Type-Specific Expert Model}

Each expert $E_j$ is associated with the $j$-th cell type and produces its expert-specific prediction through two steps: local-context refinement followed by co-expression-aware prediction. 
First, the C2CA module refines the target-cell feature $\mathbf{z}_i$ using neighboring-cell features, producing an expert-specific feature $\tilde{\mathbf{z}}_i^j$. 
Second, the CAP uses $\tilde{\mathbf{z}}_i^j$ together with a cell-type-specific gene co-expression graph to generate the expert prediction.

\noindent
{\bf Local Context Refinement with C2CA.}
C2CA refines the target-cell feature using neighboring-cell features as local contextual cues before CAP prediction. 
In tissue, cells of the same or related types often appear in local neighborhoods, and such neighboring cells can provide useful cues when the morphology of a single target cell is ambiguous. 
For each target cell, C2CA collects cell features within an $L \times L$ region around the target-cell position and uses a lightweight cross-attention module to aggregate informative neighboring features. 
The aggregated context is added to the target-cell feature and passed through a feed-forward network to obtain $\tilde{\mathbf{z}}_i^j$, which is then used as the input to CAP in the $j$-th expert.

\noindent
{\bf Cell-Type-Specific Co-Expression-Aware Predictor (CAP).}
CAP leverages prior knowledge of cell-type-specific gene co-expression patterns obtained from large-scale scRNA-seq databases, such as the Gene Expression Omnibus (GEO)~\cite{barrett2012ncbi}. 
For each expert, we extract scRNA-seq profiles corresponding to its associated cell type and compute pairwise correlations between genes. 
Gene pairs whose correlation exceeds a threshold $\tau$ are selected as strongly co-expressed pairs, forming a cell-type-specific co-expression graph $\mathcal{G}^j=(\mathcal{V}^j,\mathcal{E}^j)$, where nodes represent genes and edges connect co-expressed gene pairs.

We use this graph to construct co-expression-aware prediction parameters. 
Each gene node is assigned a learnable embedding $\mathbf{U}^j \in \mathbb{R}^{G \times d_g}$, and the graph is processed by a graph convolutional network:
\begin{equation}
\mathbf{H}^j = \mathrm{GCN}^j(\mathbf{U}^j, \mathcal{G}^j), \quad \mathbf{H}^j \in \mathbb{R}^{G \times D}.
\end{equation}
Each row of $\mathbf{H}^j$ serves as a gene-wise prediction vector. 
Given the expert-specific feature $\tilde{\mathbf{z}}_i^j$, the prediction of the $j$-th expert is obtained as
\begin{equation}
\hat{\mathbf{y}}_i^{j} = \mathbf{H}^j \tilde{\mathbf{z}}_i^j .
\end{equation}
This design encourages co-expressed genes to share similar prediction parameters, enabling each expert to reflect cell-type-specific gene-gene structure.

To preserve the co-expression prior during training, we introduce a regularization term that encourages connected genes to have similar representations:
\begin{equation}
\mathcal{L}_{\mathrm{reg}}^{j}
=
\sum_{(u,v)\in \mathcal{E}^j}
\left(1-\cos(\mathbf{h}_{u}^{j},\mathbf{h}_{v}^{j})\right),
\quad
\mathcal{L}_{\mathrm{reg}}
=
\frac{1}{C}
\sum_{j=1}^{C}
\mathcal{L}_{\mathrm{reg}}^{j},
\end{equation}
where $\mathbf{h}_{u}^{j}$ and $\mathbf{h}_{v}^{j}$ denote the rows of $\mathbf{H}^j$ corresponding to genes $u$ and $v$, respectively.

\subsubsection{Overall Loss Function}\label{sec:overall_loss}

The trainable components of GC-MoE are jointly optimized by combining the expression prediction loss, the routing loss, and the co-expression regularization:
\begin{equation}
\mathcal{L}
=
\mathcal{L}_{\mathrm{pcc}}
+
\mathcal{L}_{\mathrm{route}}
+
\lambda \mathcal{L}_{\mathrm{reg}}.
\end{equation}
Here, $\mathcal{L}_{\mathrm{pcc}}$ is backpropagated through the final aggregated prediction and optimizes both the routing model $R$ and the experts $\{E_j\}_{j=1}^{C}$, $\mathcal{L}_{\mathrm{route}}$ directly supervises only $R$, and $\mathcal{L}_{\mathrm{reg}}$ regularizes only the expert-side CAP parameters. 
The hyperparameter $\lambda$ controls the strength of the co-expression regularization.

\section{Experiments}

\noindent
{\bf Datasets and Evaluation Settings.}
We used three datasets, \textbf{10x Xenium}, \textbf{COAD}, and \textbf{IDC}, in which single-cell ST are measured using Xenium technology~\cite{janesick2023high}.
The \textbf{10x Xenium} dataset was obtained from data released by the 10x Xenium platform and consists of a breast cancer sample from a single patient, containing 5,508 cells paired with ST~\cite{fu2025spatial}.
\textbf{COAD} and \textbf{IDC} were obtained from the publicly available HEST-1k dataset~\cite{jaume2024hest} and correspond to the bowel and breast, respectively, each consisting of four pathology images. In total, they contain 830,644 and 1,735,885 cells with associated ST measurements.
We used four cell-type labels across all datasets: neoplastic, inflammatory, connective, and epithelial. For \textbf{10x Xenium}, these labels were derived from the provided annotations, while for \textbf{COAD} and \textbf{IDC}, they were assigned using a gene expression-based cell-type annotation method~\cite{nishimura2026leveraging}.
Detailed preprocessing procedures for each dataset are provided in the supplementary material.

To acquire prior knowledge of co-expression between genes, we used GEO, a publicly available repository for single-cell expression data.  
For \textbf{10x Xenium} and \textbf{IDC}, we used gene expression data from 100,064 cells obtained from breast cancer patients, as reported in~\cite{wu2021single}.  
For \textbf{COAD}, we used gene expression data from 49,859 cells obtained from colorectal cancer patients, as reported in~\cite{khaliq2022refining}.  
For each dataset, we used the same gene set as that of the target dataset to construct the co-expression prior. These datasets also provide cell-type annotations.

Performance on each dataset was evaluated using cross-validation. For the \textbf{10x Xenium} dataset, following~\cite{fu2025spatial}, we conducted 5-fold cross-validation by splitting each slide into five vertical regions. For \textbf{COAD} and \textbf{IDC}, we performed leave-one-slide-out cross-validation using four slides. The average performance on each dataset is reported as the final result.

To ensure a noise-robust evaluation in single-cell ST, we follow the original HEST-1k evaluation setting~\cite{jaume2024hest} and use the top 50 highly variable genes as prediction targets.
We evaluate model performance using the Pearson correlation coefficient (PCC)~\cite{chung2024accurate}, computed as the correlation between the ground-truth and predicted expression values for each gene and then averaged across all genes.

\noindent
{\bf Implementation Details.}
All experiments were implemented in PyTorch~\cite{Paszke2019PyTorchAI} using the Adam optimizer~\cite{adam} with a learning rate of $3 \times 10^{-5}$.
As the feature extractor $f$, we used CONCH~\cite{lu2024visual}, a large-scale pretrained foundation model for histopathological image analysis, whose parameters were frozen during training.
The proposed method was trained for up to 1,000 epochs with a mini-batch size of 1,024, and early stopping with a patience of 50 was applied.
The number of experts was set to $C=4$, corresponding to the number of cell types. We set the hyperparameters to $\tau = 0.25$, $L = 1{,}024~\mu\mathrm{m}$, and $\lambda = 0.1$.

\begin{table}[t]
    \centering
    \renewcommand{\arraystretch}{1.3} %
    \caption{\textbf{Comparison of single-cell ST estimation performance} on the 10x Xenium, COAD, and IDC datasets. Methods marked with * denote approaches originally developed for spot-level ST estimation and reimplemented for the single-cell setting. Results are reported as mean $\pm$ standard deviation over cross-validation, with the best performance in bold.}
    \scalebox{1}{
    \setlength{\tabcolsep}{6.5pt}
    \begin{tabular}{ccccc}
        \toprule
        {Method}& \textbf{10x Xenium}&\textbf{COAD}&\textbf{IDC}&\textbf{Average}\\
        \midrule
        GHIST~\cite{fu2025spatial}            &0.293 $\pm$ 0.014&0.118 $\pm$ 0.046&0.070 $\pm$ 0.045& 0.160\\
        \hdashline
        Single-Cell  ST-Net*~\cite{he2020integrating}          &0.321 $\pm$ 0.028&0.202 $\pm$ 0.061&0.166 $\pm$ 0.050&0.230 \\
        Single-Cell HisToGene*~\cite{pang2021leveraging}   &0.340 $\pm$ 0.034&0.183 $\pm$ 0.041&0.146 $\pm$ 0.040&0.223 \\
        Single-Cell  BLEEP*~\cite{xie2023spatially}          &0.312 $\pm$ 0.024&0.159 $\pm$ 0.059&0.100 $\pm$ 0.046&0.190 \\
        Single-Cell  TRIPLEX*~\cite{chung2024accurate}        &0.346 $\pm$ 0.030&0.219 $\pm$ 0.076& 0.184 $\pm$ 0.055&0.250\\
        \rowcolor{gray!15}
        \textbf{Ours} &\textbf{0.357} $\pm$ 0.028&\textbf{0.233} $\pm$ 0.055&\textbf{0.191} $\pm$ 0.039&\textbf{0.260}\\
        \bottomrule
    \end{tabular}
    }
    \label{tab:comparison}
    \vspace{-1mm}
\end{table}

\section{Comparative Experiments}

Single-cell ST prediction remains underexplored, with only a limited number of existing methods.
Therefore, we compare our method with this method, as well as with baseline methods obtained by re-implementing and adapting spot-level ST approaches to the single-cell setting.
1) ``GHIST''~\cite{fu2025spatial} is the first method for single-cell ST prediction, employing an integrated U-Net-based framework for joint cell segmentation and gene expression prediction. Because its feature extraction and prediction components are tightly coupled within the U-Net architecture, we evaluate GHIST using its original framework rather than replacing its backbone with our cell-centered foundation-model features.
2–5) are re-implemented spot-level ST methods. For all methods, we extract single-cell features using our pipeline and apply their original prediction strategies.
2) ``Single-cell ST-Net''~\cite{fu2025spatial} applies a single regression model to single-cell features.
3) ``Single-cell HisToGene''~\cite{pang2021leveraging} incorporates a transformer encoder to enable spatial information sharing across cells.
4) ``Single-cell BLEEP''~\cite{xie2023spatially} aligns single-cell features with gene expression in a shared latent space.
5) ``Single-cell TRIPLEX''~\cite{chung2024accurate} exploits spatial information through multi-scale feature aggregation.

Table~\ref{tab:comparison} presents the comparative results. 
``GHIST'' obtains the lowest performance. 
Because ``GHIST'' tightly couples feature extraction with segmentation and prediction in a U-Net-based framework, we evaluate it using its original architecture. 
Its lower performance may be partly due to the conventional backbone and single shared predictor, which may be insufficient for capturing fine-grained cellular morphology and heterogeneous single-cell expression patterns.
Among the adapted spot-level baselines, ``Single-cell ST-Net'' performs competitively with a simple regression head on cell-centered foundation-model features. 
Methods that incorporate spatial context, such as ``HisToGene'' and ``TRIPLEX'', show further improvements on some datasets, with ``TRIPLEX'' achieving the strongest baseline performance. 
In contrast, ``BLEEP'' performs worse than ``ST-Net,'' suggesting that learning a shared image-expression latent space is challenging under the high variability of single-cell targets.

GC-MoE achieves the best performance across all datasets.
The improvement over the strongest baseline is modest but consistent, supporting the benefit of explicitly modeling cell-type-structured expression variability.
By softly combining cell-type-specific experts and incorporating gene co-expression priors, GC-MoE better captures the heterogeneous nature of single-cell gene expression than single-predictor baselines.

\section{Analysis}
\noindent
{\bf Ablation Study.}
To evaluate the effectiveness of each component in the proposed method, we conduct an ablation study. ``Ours w/o CS-MoE, CAP, C2CA'' is a simple baseline in which all key components of the proposed method, namely cell-type-specific MoE, CAP, and C2CA, are removed.
``Ours w/o CAP, C2CA'' introduces the cell-type-specific MoE framework, where separate experts are constructed for each cell type and assigned via a routing model, instead of using a single shared model.
``Ours w/o C2CA'' further enhances the specialization of each expert by introducing CAP, which enables prediction while considering cell-type-specific gene co-expression relationships.

\begin{table}[t]
    \centering
    \renewcommand{\arraystretch}{1.3} %
    \caption{\textbf{Ablation study.} ``CS-MoE'': Cell-type-specific Mixture-of-Experts, ``CAP'': Co-expression-aware predictor, ``C2CA'': Cell-to-cell interaction attention. Best performance in bold.}
    \scalebox{0.84}{
    \setlength{\tabcolsep}{3pt}
    \begin{tabular}{cccccccc}
        \toprule
        Method & CS-MoE & CAP & C2CA & \textbf{10x Xenium} & \textbf{COAD} & \textbf{IDC}  & \textbf{Average} \\
        \midrule
        Ours w/o CS-MoE, CAP, C2CA & \ding{55} & \ding{55} & \ding{55} &0.321 $\pm$ 0.028&0.202 $\pm$ 0.061&0.166 $\pm$ 0.050& 0.230\\
        Ours w/o CAP, C2CA         & \ding{51} & \ding{55} & \ding{55} &0.331 $\pm$ 0.029
&0.212 $\pm$ 0.063&0.167 $\pm$ 0.049& 0.237\\
        Ours w/o C2CA               & \ding{51} & \ding{51} & \ding{55} &0.352 $\pm$ 0.026&0.224 $\pm$ 0.062&0.178 $\pm$ 0.046&0.251\\
        \rowcolor{gray!15}
        \textbf{Ours}           & \ding{51} & \ding{51} & \ding{51} &\textbf{0.357} $\pm$ 0.028&\textbf{0.233} $\pm$ 0.055&\textbf{0.191} $\pm$ 0.039& \textbf{0.260}\\
        \bottomrule
    \end{tabular}
    }
    \label{tab:ablation}
    \vspace{-2mm}
\end{table}

Table~\ref{tab:ablation} presents the ablation results. 
Adding CS-MoE improves performance over the single-predictor baseline, suggesting that cell-type-specific experts help model heterogeneous single-cell expression patterns. 
Incorporating CAP further improves performance, showing the benefit of using cell-type-specific gene co-expression priors within each expert. 
The full model with C2CA achieves the best performance, indicating that neighboring-cell context provides additional complementary information. 
These results support the effectiveness of the proposed cell-type-specific MoE and co-expression-aware expert design, with C2CA serving as a lightweight contextual refinement.

\begin{figure}[t]
      \centering
        \includegraphics[width=0.84\linewidth]{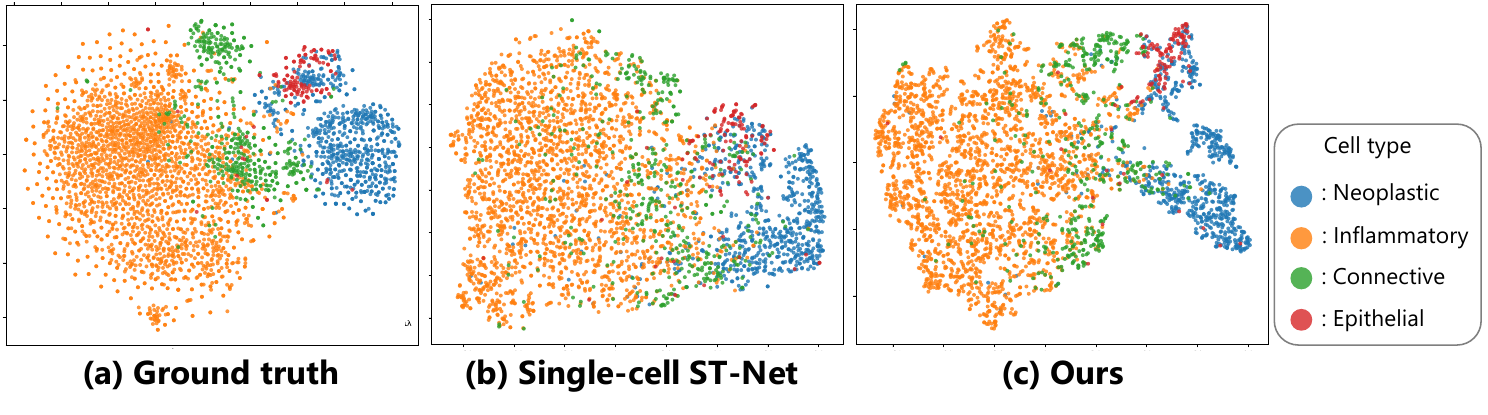}
        \vspace{-2mm}
        \caption{\textbf{Visualization of gene expression distributions.} (a) Ground-truth expression and predictions from (b) “Single-cell ST-Net” and (c) our proposed GC-MoE are visualized using t-SNE. }
        \label{fig:distribution_analysis}
        \vspace{-4mm}
\end{figure}

\noindent
{\bf Analysis of Predicted Gene Expression Distributions.}
To demonstrate that ``GC-MoE'' improves modeling of cell-type-specific expression distributions, we visualize the predicted gene expression. Figure~\ref{fig:distribution_analysis} shows t-distributed stochastic neighbor embedding (t-SNE) visualizations of (a) ground-truth expression and predictions from (b) ``Single-cell ST-Net'' and (c) GC-MoE on the 10x Xenium dataset.
The ground-truth expression forms cell type–specific cluster-like distributions. In contrast, the expression estimated by “Single-cell ST-Net,” which uses a single regression model, is less tightly clustered and more dispersed across cell types. 

The proposed method produces more distinct cell-type-specific distributions than ``Single-cell ST-Net.'' This improvement is likely due to the use of cell-type-specific expert models, enabling better modeling of complex expression patterns.

\begin{wrapfigure}{r}{0.44\textwidth}
    \centering
    \vspace{-13pt}
    \includegraphics[width=0.43\textwidth]{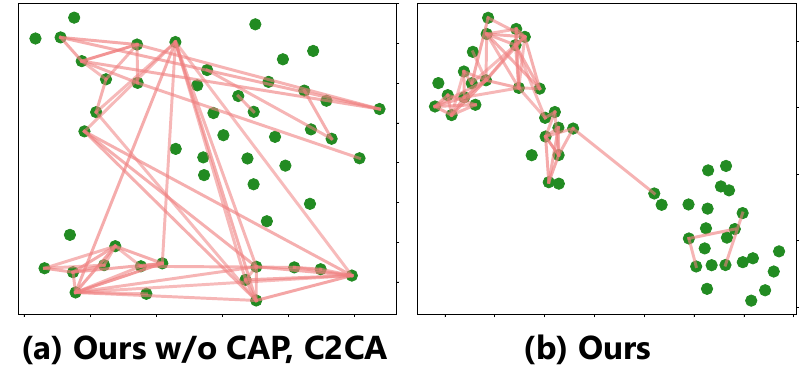}
    \vspace{-10pt}
    \caption{{\bf Visualization of CAP in GC-MoE.} t-SNE plots of the neoplastic expert’s final-layer prediction parameters for ``Ours w/o CAP, C2CA'' and ``Ours.'' Red lines indicate gene pairs with relatively strong correlations (correlation > 0.4) in the target dataset.}
    \label{fig:visualization_CAP}
        \vspace{-2mm}
\end{wrapfigure}

\noindent
{\bf Analysis of the Co-Expression-Aware Predictor.}
CAP encourages co-expressed genes to have similar prediction parameters. To examine this effect, we analyze the geometry of gene-wise prediction parameters learned by the neoplastic expert. If co-expression structure is captured, genes with strong expression correlations should be close in this parameter space; we therefore visualize the parameters using t-SNE. We compare (a) the final-layer linear parameters of ``Ours w/o CAP, C2CA'' (i.e., CS-MoE only), which does not use co-expression information, and (b) the gene-wise prediction parameters of GC-MoE with CAP.

Figure~\ref{fig:visualization_CAP} shows the results, where each point represents the prediction parameter of a gene. Red links connect gene pairs with relatively strong expression correlations (correlation $>0.4$) in the target dataset. In ``Ours w/o CAP, C2CA,'' correlated genes are relatively scattered, whereas GC-MoE places many of them closer in the embedding space. This observation is consistent with CAP encouraging co-expressed genes to share similar prediction parameters.

\begin{figure}[t]
      \centering
        \includegraphics[width=0.82\linewidth]{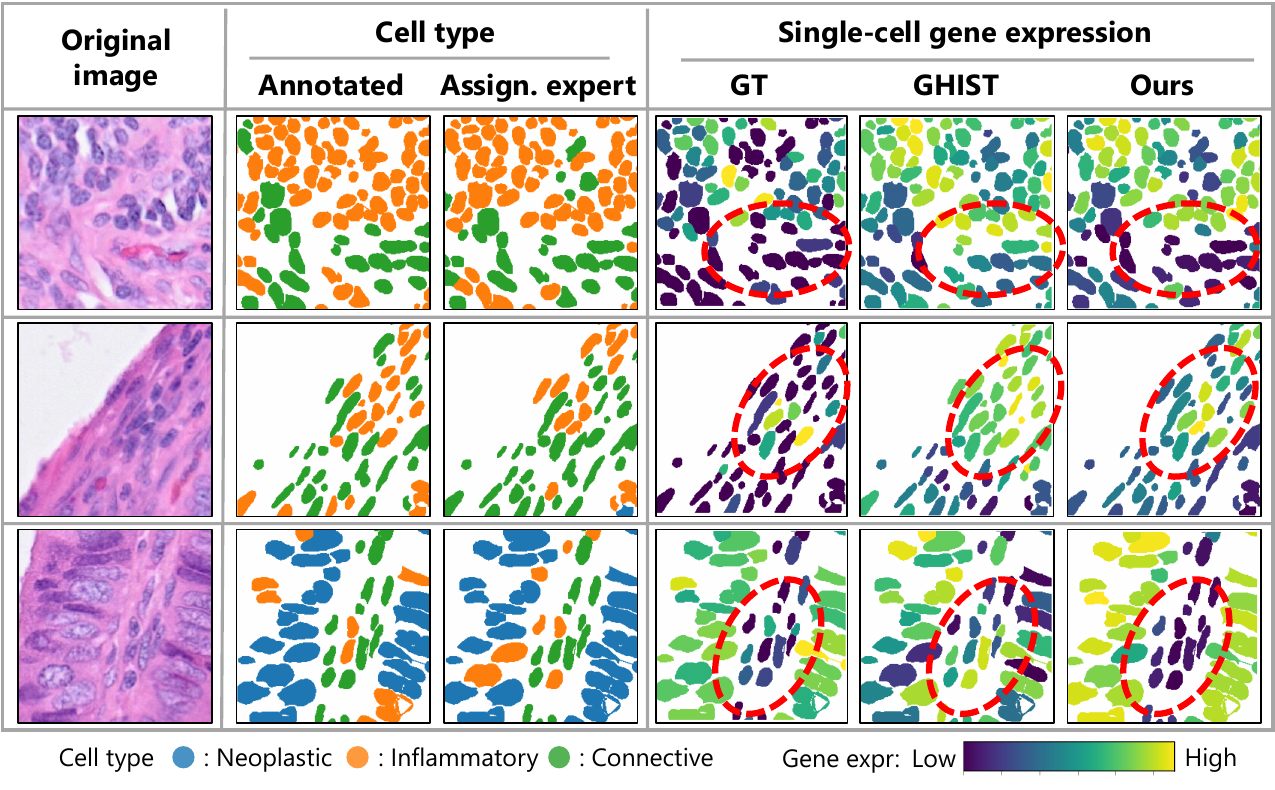}
        \vspace{-2mm}
        \caption{\textbf{Visualization of expert assignments and predicted gene expression by the proposed method.} From left to right, we show the original tissue image, the corresponding cell-type mask, per-cell expert assignments by the proposed method, and the expression levels: ground truth, those estimated by GHIST, and those estimated by our method.}
        \label{fig:vis_expr_level}
        \vspace{-4mm}
\end{figure}

\noindent
{\bf Visualization of Expert Assignments and Predicted Gene Expression by the Proposed Method.}
To qualitatively demonstrate that the proposed ``GC-MoE'' assigns appropriate cell-type-specific experts to each cell and captures cell-type-specific differences in gene expression, we visualize the predictions on tissue images from the COAD dataset.

Figure~\ref{fig:vis_expr_level} shows, from left to right, the original tissue image, the corresponding cell-type mask annotated using ground-truth gene expression, the experts assigned to each cell by the proposed method, and the gene expression levels of ground truth, the estimates produced by ``GHIST,'' and the estimates produced by our proposed ``GC-MoE'' (Ours). The gene expression visualizations were generated for multiple genes, namely IL7R, CXCR4, and CEACAM6, from top to bottom.
By comparing the ground-truth-based cell-type mask with the expert assignment mask produced by the routing model, we observe that the assigned experts generally match the annotated cell types.

Comparison between ground-truth gene expression levels and those estimated by ``GHIST'' shows that ``GHIST'' fails to accurately capture expression changes across cell types, resulting in overly smoothed predictions across different cell types.
In contrast, the proposed method more successfully captures cell-type-specific differences in gene expression.
This result suggests that the improvement is achieved by the expert assignment strategy of the proposed ``GC-MoE'', in which the routing model assigns experts corresponding to the cell type of each target cell.

\section{Conclusion and Limitations}
\noindent
{\bf Conclusion.}
In this work, we addressed the underexplored task of estimating ST at single-cell resolution from histopathology images, which is challenging due to the high diversity of gene expression patterns despite subtle morphological variations. We proposed GC-MoE, which leverages cell-type-specific expression patterns via expert models, and introduced CAP and C2CA modules to incorporate co-expression knowledge and cell–cell interactions. Experimental results showed that the proposed method consistently outperforms existing approaches and validate each component.

\noindent
{\bf Limitations.}
While we demonstrated the effectiveness of the proposed method for single-cell ST estimation, the task remains highly challenging, mainly due to severe dropout noise during observation. Compared to spot-level measurements, single-cell expression signals are weaker, making them more susceptible to missing signals and resulting in higher dropout rates. While the proposed CAP module alleviates this issue by leveraging gene co-expression patterns, it provides a partial mitigation. More explicit approaches to address dropout are needed. 
A potential direction is to incorporate more stable observational data, such as spot-level or bulk measurements, as auxiliary signals for training.

\begin{ack}
This work was supported by JST as part of ASPIRE (Grant No. JPMJAP2403) and BOOST (Grant No. JPMJBS2406).
\end{ack}

\bibliographystyle{plain}  %
\bibliography{refs}  %

\end{document}